\documentclass{article}

\usepackage{microtype}
\usepackage{graphicx}
\usepackage{subfig}
\usepackage{booktabs}
\usepackage{mathtools}
\usepackage{amsmath}
\usepackage{color}
\usepackage{caption}
\usepackage{bm}
\usepackage{bbm}
\usepackage{float}
\usepackage{amssymb}

 \usepackage{hyperref}

\usepackage[accepted]{icml2019}

\newcommand{\bx}{\bm{x}}
\newcommand{\by}{\bm{y}}

\newcommand{\BX}{\mathbf{X}}
\newcommand{\BZ}{\mathbf{Z}}

\icmltitlerunning{Bayesian Graph Convolutional Neural Networks using Node Copying}

\begin{document}
\twocolumn[
\icmltitle{Bayesian Graph Convolutional Neural Networks using Node Copying}



\icmlsetsymbol{equal}{*}
\vspace{-0.5cm}
\begin{icmlauthorlist}
\icmlauthor{Soumyasundar Pal}{equal,mcgill}
\icmlauthor{Florence Regol}{equal,mcgill}
\icmlauthor{Mark Coates}{mcgill}
\end{icmlauthorlist}

\icmlaffiliation{mcgill}{Dept. of Electrical and Computer Engineering, McGill University, Montr\'eal, Canada}

\icmlcorrespondingauthor{Soumyasundar Pal}{soumyasundar.pal@mail.mcgill.ca}

\icmlkeywords{Bayesian GCNN, Monte Carlo}

\vskip 0.1in
]



\printAffiliationsAndNotice{\icmlEqualContribution} 

\begin{abstract}
Graph convolutional neural networks (GCNN) have numerous
applications in different graph based learning tasks. Although the techniques obtain impressive results, they often fall short in accounting for the uncertainty associated with the underlying graph structure. In the recently proposed Bayesian GCNN (BGCN) framework, this issue is tackled by viewing the observed graph as a sample from a parametric random graph model and targeting joint inference of the graph and the GCNN weights. In this paper, we introduce an alternative generative model for graphs based on copying nodes and incorporate it within the BGCN framework. Our approach has the benefit that it uses information provided by the node features and training labels in the graph topology inference. Experiments show that the proposed algorithm compares favourably to the state-of-the-art in benchmark node classification tasks.
\end{abstract}

\vspace{-0.75cm}
\section{Introduction}
\label{sec:intro}
\vspace{-0.125cm}

Recently, there has been an increased research focus on graph convolutional neural networks (GCNNs) due to their successful application in various graph based learning problems such as node
and graph classification, matrix completion, and learning of node
embeddings. Prior work leading to the development of GCNNs includes~\cite{bruna2013,henaff2015,duvenaud2015}. ~\cite{defferrard2016} propose an approach based on spectral filtering which is also followed in~\cite{levie2017,chen2018,kipf2017}. Other works~\cite{atwood2016,hamilton2017b} consider spatial filtering and aggregation strategies. A general framework for learning on graphs and manifolds with neural networks is derived in~\cite{monti2017} and this includes various other existing methods as special cases.

Several modifications can improve the performance of the GCNN, including adding attention nodes~\cite{velivckovic2018}, gates~\cite{li2016b,bresson2017}, edge
conditioning and skip connections~\cite{sukhbaatar2016,simonovsky2017}. Other approaches involve the use of graph ensembles~\cite{anirudh2017}, multiple adjacency matrices~\cite{such2017}, the dual graph~\cite{monti2018}, or random perturbation~\cite{sun2019}. Employing localized sampling methods~\cite{hamilton2017b}, importance sampling~\cite{chen2018} or control variate based stochastic approximation~\cite{chen2018c} has been shown to  improve the scalability of these methods for processing large graphs.

The majority of the existing approaches process the graph as the ground truth. However, in many practical settings, the graph is often derived from noisy data or
inaccurate modelling assumptions. As a result, spurious edges may be present or 
edges between very similar nodes might be omitted. This can lead to deterioration in the performance of the learning algorithms. Various existing approaches like the graph attention network~\cite{velivckovic2018} and graph ensemble based approach~\cite{anirudh2017} address this issue partially. Nevertheless, neither of these methods has the flexibility to add edges that could be missing from the observed graph. A principled way to address the uncertainty in the graph structure is to consider the graph as a random sample drawn from a probability distribution over graphs. The Bayesian framework of~\cite{zhang2019} proposes to use a parametric random graph model as the generative model of the graph and formulates the learning task as the inference of the joint posterior distribution of the graph and the weights of the GCNN. Despite the effectiveness of the approach, the choice of a suitable random graph model is crucial and heavily dependent on the learning task and datasets. Furthermore, the method in ~\cite{zhang2019} conducts the posterior inference of the graph solely conditioned on the observed graph topology. This results in a complete disregard of any information provided by the node features and the training labels, which is undesirable if these data are highly correlated with the true graph structure. 

In this paper, we introduce a novel generative model for graphs based on copying nodes from one location to another. While this idea is similar to the full duplication process presented in~\cite{chung2003}, we do not grow the graph since we only copy existing nodes rather than adding new ones. This results in a formulation in which the posterior inference of the graph is carried out conditioned on the features and training labels as well as the observed graph topology. Experimental results demonstrate the efficacy of our approach for the semi-supervised node classification task, particularly if a limited number of training labels is available. The rest of the paper is organized as follows. We provide a brief review of the GCNN in
Section~\ref{sec:gcnn} and present the proposed approach in
Section~\ref{sec:methodology}. We report the results of the numerical experiments
in Section~\ref{sec:experimental_results} and make concluding remarks in Section~\ref{sec:conclusion}.
\vspace{-0.125cm}
\section{Graph convolutional neural networks}
\label{sec:gcnn}
\vspace{-0.125cm}
Although graph convolutional neural networks are suitable for a
variety of learning tasks, here we restrict ourselves to the discussion of the node classification problem on a graph for brevity. In this setting, 
an observed graph $\mathcal{G}_{obs} = (\mathcal{V},\mathcal{E})$ is available,
where $\mathcal{V}$ is the set of $N$ nodes and $\mathcal{E} \subseteq \mathcal{V} \times \mathcal{V}$ denotes
the set of edges. There is a feature vector $\bx_i \in \mathbf{R}^{d \times 1}$ associated with each node $i$ and its class label is denoted by $\by_i$. The labels are known only for the nodes in the training set $\mathcal{L}\subset \mathcal{V}$. The goal is to
predict the labels of the remaining nodes using the information provided
by the observed graph $\mathcal{G}_{obs}$, the feature matrix
$\BX = [\bx_1, \bx_2, \dots, \bx_N]^T$ and the training labels
$\mathbf{Y_{\mathcal{L}}} = \{\by_i: i \in \mathcal{L}\}$.


In a GCNN, learning is performed using graph convolution operations within a neural network architecture. A layerwise propagation rule for the simpler architectures~\cite{defferrard2016,kipf2017} is written as:

\begin{align}
\mathbf{H}^{(1)} &=\sigma(\mathbf{\hat{A}}_{\mathcal{G}}\mathbf{X}\mathbf{W}^{(0)}) \,,\\
\mathbf{H}^{(l+1)} &= \sigma(\mathbf{\hat{A}}_{\mathcal{G}}\mathbf{H}^{(l)}\mathbf{W}^{(l)})\,. 
\end{align}
The normalized adjacency operator $\mathbf{\hat{A}}_{\mathcal{G}}$ is derived from the observed graph and it controls the aggregation of the output features across the neighbouring nodes at each layer. $\sigma$ denotes a pointwise non-linear activation function and $\mathbf{H}^{(l)}$ are the output features from layer $l-1$. $\mathbf{W}^{(l)}$ represents the weights of the neural network at layer
$l$. We use $\mathbf{W}=\{\mathbf{W}^{{l}}\}_{l=1}^L$ to denote the collection of GCNN weights across all layers. In an $L$-layer network, the final output is collected from the last layer $\mathbf{Z} =
\mathbf{H}^{(L)}$. The weights of the neural network $\mathbf{W}$ are learned via
backpropagation with the objective of minimizing an error metric between
the training labels $\mathbf{Y_{\mathcal{L}}}$ and the network predictions $\mathbf{Z_{\mathcal{L}}} = \{\bm{z}_i: i \in \mathcal{L}\}$ at the nodes in the training set.
\vspace{-0.125cm}
\section{Methodology}
\label{sec:methodology}
\vspace{-0.125cm}
In the Bayesian paradigm, the observed graph is viewed as a random quantity and the posterior inference for the underlying graph is required.
We postulate a model which allows sampling of a random graph by copying the observed graph and then replacing each node's edges with a high probability by the edges of a similar node randomly selected from the observed graph, while the node features remain unchanged.
\subsection{Node-Copying Graph Model}
\label{subsec:copying_model}
\vspace{-0.125cm}
In order to sample graph $\mathcal{G}$ from the proposed model, we introduce an auxiliary random vector $\boldsymbol{\zeta} \in \{1,2,... N\}^N$, where the $j$'th entry $\zeta^j$  denotes the node whose edges are to replace the edges of the $j$'th node in the observed graph. The entries in $\boldsymbol{\zeta}$ are assumed to be mutually independent. For sampling the $\zeta^j$s, we use a base classification algorithm using the observed graph $\mathcal{G}_{obs}$, the features $\BX$ and the training labels $\mathbf{Y_{\mathcal{L}}}$ to obtain labels $\hat{c}_{\ell} \in \{1,2,...K\}$ for each node $\ell$ in the graph. Then for each class $1 \leq k \leq K$,  we collect the nodes with predicted label $k$ into the set $\mathcal{C}_k$:
\begin{align}
    \mathcal{C}_k = \{\ell \mid 1 \leq \ell \leq N, \hat{c}_{\ell} = k\}\,.\label{eq:C_k}
\end{align}
We define the posterior distribution of $\boldsymbol{\zeta}$  as follows:
 \begin{align}
 p(\boldsymbol{\zeta}|\mathcal{G}_{obs}, \BX, \mathbf{Y_{\mathcal{L}}}) &= \prod_{j=1}^{N}p(\zeta^j|\mathcal{G}_{obs}, \BX, \mathbf{Y_{\mathcal{L}}})\,,\nonumber\\
p(\zeta^j = m|\mathcal{G}_{obs}, \BX, \mathbf{Y_{\mathcal{L}}}) &= \begin{dcases} \frac{1}{|\mathcal{C}_k|}, \enspace \text{if } \hat{c}_j = \hat{c}_m = k\,\\
0, \enspace  \text{      otherwise}\label{eq:dist_zeta}\,,
\end{dcases}
 \end{align}
for $1 \leq j,m \leq N$ and $1 \leq k \leq K$. Sampling $\zeta^j$ from this model boils down to selecting a node at random from the collection of nodes that have the same predictive label as the $j$'th node. Conditioned on $\boldsymbol{\zeta}$ and the observed graph $\mathcal{G}_{obs}$, the sampling of graph $\mathcal{G}$ is carried out by copying the $\zeta^j$'th node of $\mathcal{G}_{obs}$ in the place of the $j$'th node of $\mathcal{G}$, independently for all $1 \leq j \leq N$ with a high probability. More formally, the generative model is given as:
\begin{align}
\hspace{-0.3cm}p(\mathcal{G}|\mathcal{G}_{obs}, \boldsymbol{\zeta}) = \prod_{j=1}^N \epsilon^{\mathbbm{1}_{\{\mathcal{G}_j = \mathcal{G}_{obs,j}\}}}(1 - \epsilon)^{\mathbbm{1}_{\{\mathcal{G}_j = \mathcal{G}_{obs,\zeta^j}\}}}\label{eq:copying_model}\,,
\end{align}
where, $0<\epsilon \ll 1$ is a hyperparameter and $\mathbbm{1}_{\{\mathcal{G}_j = \mathcal{G}_{obs,q}\}}$ denotes the indicator function of copying $q$'th node of $\mathcal{G}_{obs}$ in place of the $j$'th node of $\mathcal{G}$. The copying operation involves changing the set of neighbours of the $j$'th node of $\mathcal{G}$ to be the same as the set of neighbours of the $q$'th node of $\mathcal{G}_{obs}$.

\subsection{Bayesian Graph Convolutional Neural Networks}
\label{subsec:bgcn}
\vspace{-0.125cm}
As in~\cite{zhang2017}, we compute the marginal posterior probability of the node labels via marginalization with respect to the graph and the GCNN weights.
\begin{align}
p(\BZ|\mathbf{Y_{\mathcal{L}}},&\BX,\mathcal{G}_{obs}) 
= \int p(\BZ|\mathbf{W},\mathcal{G}_{obs},\BX) p(\mathbf{W}|\mathbf{Y_{\mathcal{L}}},\BX,\mathcal{G}) \nonumber\\
&  p(\mathcal{G}|\mathcal{G}_{obs}, \boldsymbol{\zeta})p(\boldsymbol{\zeta}|\mathcal{G}_{obs},\mathbf{Y_{\mathcal{L}}},\BX) \,d\mathbf{W}\,d\mathcal{G}\,d\boldsymbol{\zeta}\label{eq:exact_posterior}\,.
\end{align}
Here $\mathbf{W}$ denotes the random weights of a Bayesian GCNN over
the graph $\mathcal{G}$ and $\boldsymbol{\zeta}$ is an $N$-dimensional random vector associated with the proposed node copying model. In a node classification problem with $K$ classes, the term
$p(\BZ|\mathbf{Y_{\mathcal{L}}},\BX,\mathcal{G}_{obs})$ is modelled
using a $K$-dimensional categorical distribution by applying a softmax function to the
output of the GCNN. In \citep{zhang2019}, $\mathcal{G}_{obs}$
is viewed as a sample realization from a collection of graphs associated with a parametric random graph model and posterior inference of
$p(\mathcal{G}|\mathcal{G}_{obs})$ is targeted via marginalization of the random
graph parameters. Their approach thus ignores any possible dependence of the graph
$\mathcal{G}$ on the features $\BX$ and the labels
$\mathbf{Y_{\mathcal{L}}}$. By contrast, our approach models the marginal posterior distribution of the graph $\mathcal{G}$ as
$p(\mathcal{G}|\mathcal{G}_{obs}, \BX, \mathbf{Y_{\mathcal{L}}})$. This allows us to incorporate the information provided by the features $\BX$ and the training labels $\mathbf{Y_{\mathcal{L}}}$ in the graph inference process. The integral in equation~\eqref{eq:exact_posterior} is not analytically tractable. Hence, a Monte Carlo approximation is formed as follows:
\begin{align}
& p(\BZ|\mathbf{Y_{\mathcal{L}}},\BX,\mathcal{G}_{obs}) \approx \nonumber \\
& \quad \quad \dfrac{1}{V} \sum_{v=1}^V
 \dfrac{1}{N_G S}\sum_{i=1}^{N_G}\sum_{s=1}^S p(\BZ|\mathbf{W}_{s,i,v},\mathcal{G}_{obs},\BX)\,.
\label{eq:MC_posterior}
\end{align}
In this approximation, $V$ samples $\boldsymbol{\zeta}_v$ are drawn from $p(\boldsymbol{\zeta}|\mathcal{G}_{obs},\mathbf{Y_{\mathcal{L}}},\BX)$. The $N_G$ graphs $\mathcal{G}_{i,v}$ are sampled from $p(\mathcal{G}|\mathcal{G}_{obs}, \boldsymbol{\zeta}_v)$ and subsequently $S$ weight matrices $\mathbf{W}_{s,i,v}$ are sampled from $p(\mathbf{W}|\mathbf{Y_{\mathcal{L}}},\BX,\mathcal{G}_{i,v})$ from the Bayesian GCN corresponding to the graph $\mathcal{G}_{i,v}$. 

\begin{algorithm}[tb]
\caption{Bayesian GCN with node copying}
\label{alg:bgcn_copying}
\begin{algorithmic}[1]
\STATE {\bfseries Input:}  $\mathcal{G}_{obs}$, $\BX$, $\mathbf{Y_{\mathcal{L}}}$
\STATE {\bfseries Output:}  $p(\BZ|\mathbf{Y_{\mathcal{L}}},\BX,\mathcal{G}_{obs})$
\STATE Initialization: train a base classifier to obtain $\hat{c}_{\ell}$ for $1 \leq \ell \leq N$, form $\mathcal{C}_k$ using eq.~\eqref{eq:C_k} for $1 \leq k \leq K$. 
\FOR{$v=1$ {\bfseries to} $V$}
\STATE Sample $\boldsymbol{\zeta}_v \sim p(\boldsymbol{\zeta}|\mathcal{G}_{obs}, \BX, \mathbf{Y_{\mathcal{L}}})$ using eq.~\eqref{eq:dist_zeta}.
\FOR{$i=1$ {\bfseries to} $N_G$}
\STATE Sample graph $\mathcal{G}_{i,v} \sim p(\mathcal{G}|\mathcal{G}_{obs}, \boldsymbol{\zeta}_v)$ using eq.~\eqref{eq:copying_model}.
\FOR{$s=1$ {\bfseries to} $S$}
\STATE Sample weights $W_{s,i,v}$ using MC dropout by training a GCNN over the graph $\mathcal{G}_{i,v}$.
\ENDFOR
\ENDFOR
\ENDFOR
\STATE Approximate $p(\BZ|\mathbf{Y_{\mathcal{L}}},\BX,\mathcal{G}_{obs})$ using eq.~\eqref{eq:MC_posterior}.
\end{algorithmic}
\end{algorithm}

Sampling graphs from the node-copying model in Section~\ref{subsec:copying_model} has several advantages compared to the graph inference technique based on mixed membership stochastic block models (MMSBMs)~\cite{airoldi2009}, which was adopted in~\cite{zhang2019}. First, the sampling of $\boldsymbol{\zeta}$ is computationally much cheaper than the inference of parameters of the parametric model, which becomes more severe as the size of the graph increases. Second, it is in general extremely difficult to carry out accurate inference for high dimensional MMSBM parameters~\cite{li2016c} and inaccuracies in parameter estimates results in sampling of graphs which are very different from the observed graph. This can impact classification performance adversely, particularly if the observed graph does not fit the MMSBM well. However, for the proposed copying model, the similarity between the sampled graph and the observed graph depends mostly on the performance of the base classifier. If a state-of-the-art graph based classification method (e.g., GCNN) is used, we can obtain more representative graph samples from this model, particularly for large graphs. The expected graph edit distance between the random graphs and the observed graph can be controlled by the choice of the parameter $\epsilon$. A low value of $\epsilon$ is chosen since it causes high variability among the random graph samples which was found to be effective empirically.
Third, sampling a graph from the MMSBM scales as $\mathcal{O}(N^2)$ whereas the proposed method offers $\mathcal{O}(N)$ complexity.

For the Bayesian inference of GCNN weights, we can use various techniques including expectation propagation~\cite{hernandez2015}, variational inference~\cite{gal2016,sun2017,louizos2017}, and Markov Chain Monte Carlo methods~\cite{neal1993,korattikara2015,li2016}. Similar to~\cite{zhang2019}, we train a GCNN on $\mathcal{G}_{i,v}$ and use Monte Carlo dropout~\cite{gal2016} to sample $\mathbf{W}_{s,i,v}$. This is equivalent to sampling the weights from a variational approximation of $p(\mathbf{W}|\mathbf{Y_{\mathcal{L}}},\BX,\mathcal{G}_{i,v})$, with a particular structure. The resulting algorithm is summarized in Algorithm~\ref{alg:bgcn_copying}.
\vspace{-0.125cm}
\section{Numerical Experiments and Results}
 \label{sec:experimental_results}
 \vspace{-0.125cm}
 We address a semi-supervised node classification task for three citation networks~\cite{sen2008}: Cora, CiteSeer, and Pubmed. In these datasets each node represents a scientific publication and an undirected edge is formed between two nodes if any one of them cites the other. Each node has a sparse bag-of-words feature vector and the label describes the topic of the document. During training, we have access to the labels of only a few nodes per class and the goal is to infer labels for the other nodes.
 
 We consider two different strategies for splitting the data into training and test sets, as specified in~\cite{zhang2019}. In the first setting, we use the fixed split from~\cite{yang2016}, which contains 20 labels per class in the training set. For the cases with 5 and 10 training labels per class in the fixed split scenario, the first 5 and 10 labels in the original partition of~\cite{yang2016} are used. The second type of split is constructed by sampling the training and test sets randomly for each trial. Since a specific split of data can impact the classification performance significantly, random splitting provides a more robust comparison of performance of the algorithms. 
 
 We compare the proposed BGCN in this paper with ChebyNet~\cite{defferrard2016}, GCNN~\cite{kipf2017}, GAT~\cite{velivckovic2018} and the BGCN in~\cite{zhang2019}. The hyperparameters of GCNN are set according to~\cite{kipf2017} and the same values are used for the BGCN algorithms as well. For the proposed BGCN, we use GCNN~\cite{kipf2017} as the base classification method. For both splitting strategies, each algorithm is run for 50 trials with random weight initializations. The average accuracies for Cora, Citeseer and Pubmed datasets along with their standard errors are reported  in Table~\ref{table:cora},~\ref{table:citeseer} and ~\ref{table:pubmed} respectively.
 \vspace{-0.125cm}
 \begin{table}[h]
	\centering
	\footnotesize{
		\begin{tabular}{lcccc}
			\toprule[0.25ex]
			\textbf{Random split} &\textbf{5 labels}        & \textbf{10 labels}         & \textbf{20 labels} \\ 
			\midrule
			\textbf{ChebyNet}             &61.7$\pm$6.8            &72.5$\pm$3.4              &78.8$\pm$1.6        \\
			\textbf{GCNN}             &70.0$\pm$3.7            & 76.0$\pm$2.2            &79.8$\pm$1.8  \\
			\textbf{GAT}              &70.4$\pm$3.7            &76.6$\pm$2.8              &79.9$\pm$1.8   \\
			\textbf{BGCN}     &\textbf{74.6$\pm$2.8}   &77.5$\pm$2.6    &80.2$\pm$1.5 \\
			\textbf{BGCN (ours)}     &73.8$\pm$2.7   &\textbf{77.6$\pm$2.6}     &\textbf{80.3$\pm$1.6} \\
			\midrule
			\textbf{Fixed split}  &\textbf{}                 & \textbf{}                  & \textbf{}\\
			\midrule
			\textbf{ChebyNet}             &67.9$\pm$3.1            &72.7$\pm$2.4             &80.4$\pm$0.7        \\
			\textbf{GCNN}             &74.4$\pm$0.8            &74.9$\pm$0.7              &\textbf{81.6$\pm$0.5}       \\
			\textbf{GAT}              &73.5$\pm$2.2            &74.5$\pm$1.3              &81.6$\pm$0.9  \\
			\textbf{BGCN}     &\textbf{75.3$\pm$0.8}   &76.6$\pm$0.8    &81.2$\pm$0.8     \\
			\textbf{BGCN (ours)}     &75.1$\pm$1.3   &\textbf{76.7$\pm$0.7}    &81.4$\pm$0.6     \\
			\bottomrule[0.25ex]
		\end{tabular}
	}
	\vspace{-0.125cm}
	\caption{Classification accuracy (in $\%$) for Cora dataset.}
	\label{table:cora}
\end{table}
\vspace{-0.5cm}
\begin{table}[h]
	\centering
	\footnotesize{
		\begin{tabular}{lcccc}
			\toprule[0.25ex]
			\textbf{Random split}  &\textbf{5 labels}        & \textbf{10 labels}         & \textbf{20 labels} \\ 
			\midrule
			\textbf{ChebyNet}             &58.5$\pm$4.8            &65.8$\pm$2.8              &67.5$\pm$1.9             \\
			\textbf{GCNN}             &58.5$\pm$4.7            &65.4$\pm$2.6              &67.8$\pm$2.3        \\
			\textbf{GAT}              &56.7$\pm$5.1            &64.1$\pm$3.3              &67.6$\pm$2.3       \\
			\textbf{BGCN}     &63.0$\pm$4.8   &\textbf{69.9$\pm$2.3}     &\textbf{71.1$\pm$1.8}   \\
			\textbf{BGCN (ours)}     &\textbf{63.9$\pm$4.2}   &68.5$\pm$2.3     &70.2$\pm$2.0   \\
			\midrule
			\textbf{Fixed split}  &\textbf{}                 & \textbf{}                  & \textbf{}\\
			\midrule
			\textbf{ChebyNet}             &53.0$\pm$1.9            &67.7$\pm$1.2              &70.2$\pm$0.9        \\
			\textbf{GCNN}             &55.4$\pm$1.1            &65.8$\pm$1.1              &70.8$\pm$0.7       \\
			\textbf{GAT}              &55.4$\pm$2.6            &66.1$\pm$1.7              &70.8$\pm$1.0    \\
			\textbf{BGCN}     &57.3$\pm$0.8   &\textbf{70.8$\pm$0.6}     &\textbf{72.2$\pm$0.6}\\
			\textbf{BGCN (ours)}     &\textbf{61.4$\pm$2.3}   &69.6$\pm$0.6     &71.9$\pm$0.6\\
			\bottomrule[0.25ex]
		\end{tabular}
	}
	\vspace{-0.125cm}
	\caption{Classification accuracy (in $\%$) for Citeseer dataset.}
	\label{table:citeseer}
\end{table}
\vspace{-0.5cm}
\begin{table}[h!]
	\centering
	\footnotesize{
		\begin{tabular}{lcccc}
			\toprule[0.25ex]
			\textbf{Random split}  &\textbf{5 labels}        &\textbf{10 labels}         &\textbf{20 labels} \\ 
			\midrule
			\textbf{ChebyNet}             &62.7$\pm$6.9            &68.6$\pm$5.0              &74.3$\pm$3.0             \\
			\textbf{GCNN}              &69.7$\pm$4.5           &73.9$\pm$3.4    &77.5$\pm$2.5  \\
			\textbf{GAT}               &68.0$\pm$4.8           &72.6$\pm$3.6             &76.4$\pm$3.0\\
			\textbf{BGCN}     &70.2$\pm$4.5  &73.3$\pm$3.1             &76.0$\pm$2.6  \\
			\textbf{BGCN (ours)}     &\textbf{71.0$\pm$4.2}  &\textbf{74.6$\pm$3.3}             &\textbf{77.5$\pm$2.4}  \\
			\midrule
			\textbf{Fixed split}  &\textbf{}                 & \textbf{}                 &\textbf{}\\
			\midrule
			\textbf{ChebyNet}             &68.1$\pm$2.5            &69.4$\pm$1.6              &76.0$\pm$1.2             \\
			\textbf{GCNN}             &69.7$\pm$0.5            &72.8$\pm$0.5            &78.9$\pm$0.3  \\
			\textbf{GAT}              &70.0$\pm$0.6   & 71.6$\pm$0.9            &76.9$\pm$0.5 \\
			\textbf{BGCN}    &70.9$\pm$0.8           &72.3$\pm$0.8    &76.6$\pm$0.7\\
			\textbf{BGCN (ours)}    &\textbf{71.2$\pm$0.5}           &\textbf{73.6$\pm$0.5}    &\textbf{79.1$\pm$0.4}\\
			\bottomrule[0.25ex]
		\end{tabular}
	}
	\vspace{-0.125cm}
	\caption{Classification accuracy (in $\%$) for Pubmed dataset.}
	\label{table:pubmed}
\end{table}

\begin{figure}[h!]
\centering
\subfloat[]{
	\label{subfig:cora}
	\vspace{-1.25cm}
	\includegraphics[scale=0.5, trim={0.5cm 0.25cm 1.5cm 1.25cm}, clip]{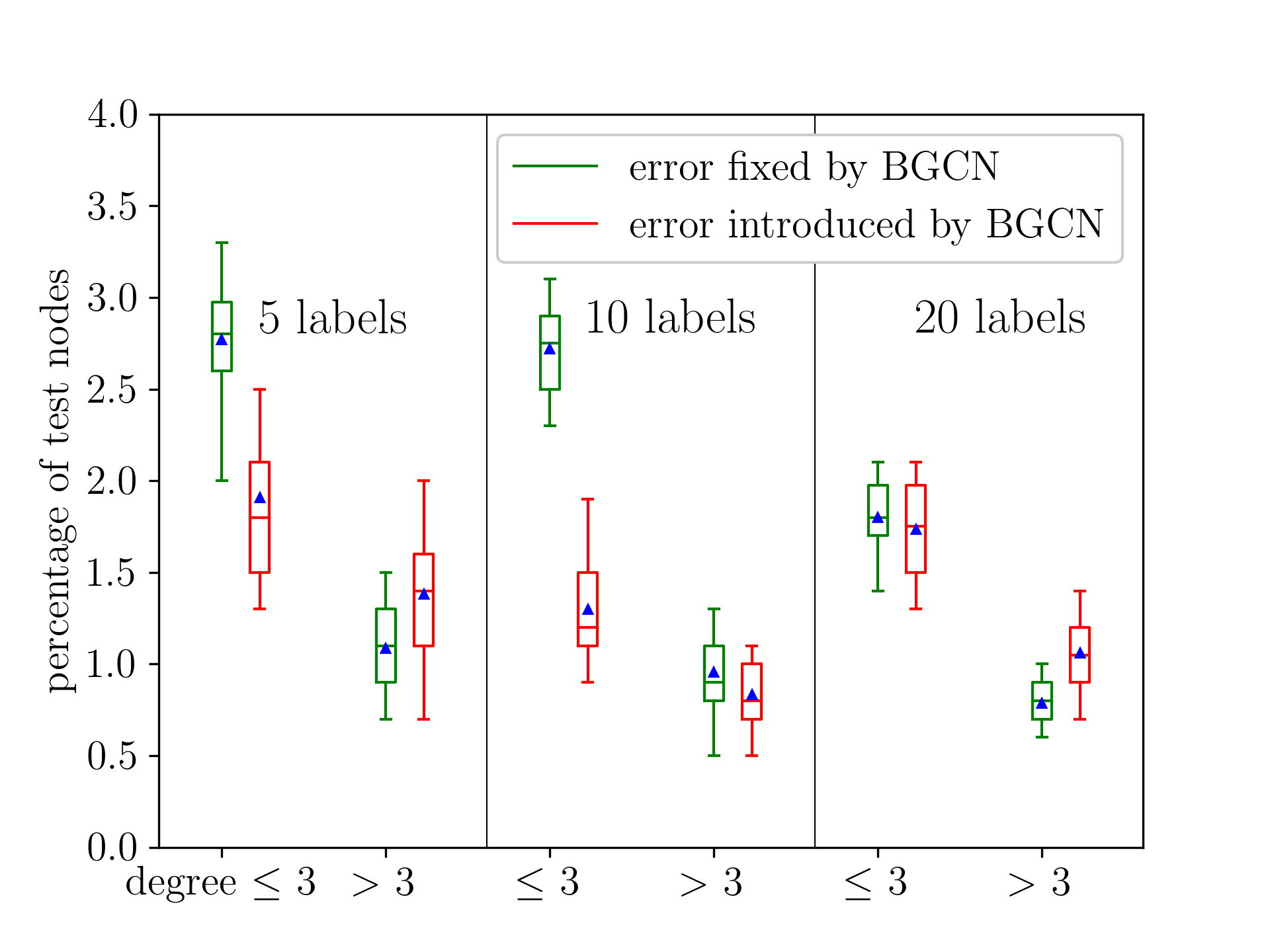}} 
 \vspace{-0.25cm}
\subfloat[]{
	\label{subfig:citeseer}
	\includegraphics[scale=0.5, trim={0.5cm 0.25cm 1.5cm 1.25cm}, clip]{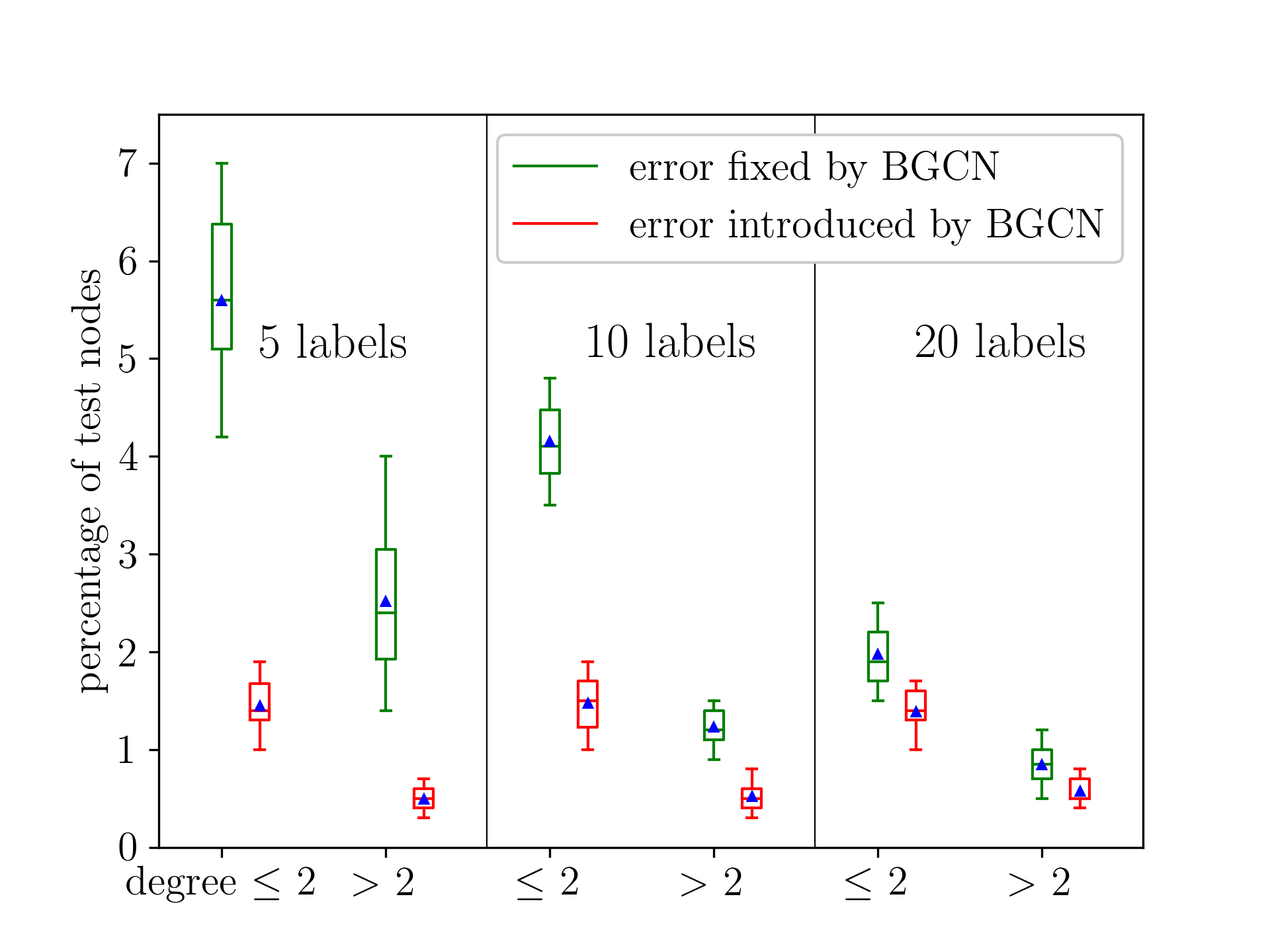}} 
 \vspace{-0.25cm}
\caption{Boxplot of different categories of nodes in the (a) Cora and (b) Citeseer datasets based on the classification results of the GCNN and the proposed BGCN algorithms. The two groups are formed by thresholding the degree of the nodes in the test set at the median value. The  box  shows  25-75  percentiles;  the  triangle represents  the  mean  value;  and  the  median  is  indicated  by  a horizontal line. Whiskers are drawn at the 5 and 95 percentiles of data points.} 
\label{fig:degree_analysis}
 
\end{figure}

We observe that the proposed BGCN algorithm obtains higher classification accuracy compared to its competitors in most cases. The improvement in accuracy compared to GCNN is more significant when the number of available labels is limited to 5 or 10. From Figure~\ref{fig:degree_analysis}, we observe that in most cases, for the Cora and the Citeseer datasets, the proposed BGCN algorithm corrects more errors of the GCNN base classifier for nodes with lower degree. 
\vspace{-0.125cm}

\vspace{-0.5cm}
\section{Conclusion}
\label{sec:conclusion}
\vspace{-0.125cm}
In this paper, we present a Bayesian GCNN using a node copying based generative model for graph. The proposed algorithm exhibits superior performance in the semi-supervised node classification task when the amount of available labels for training is limited. Future work will involve conducting a more thorough experimental evaluation and exploring ways to extend the methodology to other graph based learning tasks.

\bibliography{references}
\bibliographystyle{icml2019}

\end{document}